\documentclass[10pt,conference,a4paper]{IEEEtran}
\IEEEoverridecommandlockouts
\usepackage{cite}
\usepackage{amsmath,amssymb,amsfonts}
\usepackage{algorithmic}
\usepackage[pdftex]{graphicx}
\usepackage{tabularx}
\usepackage{textcomp}
\usepackage{xcolor}
\usepackage{subcaption}
\usepackage{url}
\usepackage{hyperref}

\def\BibTeX{{\rm B\kern-.05em{\sc i\kern-.025em b}\kern-.08em
    T\kern-.1667em\lower.7ex\hbox{E}\kern-.125emX}}
\begin{document}

\title{Low Dimensional State Representation Learning with Reward-shaped Priors}

\author{\IEEEauthorblockN{1\textsuperscript{st} Nicolò Botteghi}
\IEEEauthorblockA{\textit{Robotics and Mechatronics}\\
%\textit{Electrical Engineering, Mathematics and Computer Science} \\
\textit{University of Twente}\\
Enschede, The Netherlands \\
n.botteghi@utwente.nl}
\and
\IEEEauthorblockN{2\textsuperscript{nd} Ruben Obbink}
\IEEEauthorblockA{\textit{Robotics and Mechatronics} \\
\textit{University of Twente}\\
Enschede, The Netherlands} 
\and
\IEEEauthorblockN{3\textsuperscript{rd} Daan Geijs}
\IEEEauthorblockA{\textit{Robotics and Mechatronics} \\
\textit{University of Twente}\\
Enschede, The Netherlands} 
\and
\IEEEauthorblockN{4\textsuperscript{th} Mannes Poel}
\IEEEauthorblockA{\textit{Datamanagement and Biometrics} \\
\textit{University of Twente}\\
Enschede, The Netherlands \\
m.poel@utwente.nl}
\and
\IEEEauthorblockN{5\textsuperscript{th} Beril Sirmacek}
\IEEEauthorblockA{\textit{J\"{o}nk\"{o}ping AI Lab} \\
\textit{J\"{o}nk\"{o}ping University}\\
J\"{o}nk\"{o}ping, Sweden \\
beril.sirmacek@ju.se}
\and
\IEEEauthorblockN{6\textsuperscript{th} Christoph Brune}
\IEEEauthorblockA{\textit{Applied Analysis} \\
\textit{University of Twente}\\
Enschede, The Netherlands \\
c.brune@utwente.nl}
\and
\IEEEauthorblockN{7\textsuperscript{th} Abeje Mersha}
\IEEEauthorblockA{\textit{Research Group of Mechatronics} \\
\textit{Saxion University of Applied Sciences}\\
Enschede, The Netherlands \\
a.y.mersha@saxion.nl}
\and
\IEEEauthorblockN{8\textsuperscript{th} Stefano Stramigioli}
\IEEEauthorblockA{\textit{Robotics and Mechatronics} \\
\textit{University of Twente}\\
Enschede, The Netherlands \\
s.stramigioli@utwente.nl}
}

\maketitle

%%%%%%%%%%%%%%%%%%%%%%%%%%%%%%%%%%%%%%%%%%%%%%%%%%%%%%%%%%%%%%%%%%%%%%%%%%%%%%%%
\begin{abstract}
Reinforcement Learning has been able to solve many complicated robotics tasks without any need for feature engineering in an end-to-end fashion. However, learning the optimal policy directly from the sensory inputs, i.e the observations, often requires processing and storage of a huge amount of data. In the context of robotics, the cost of data from real robotics hardware is usually very high, thus solutions that achieve high sample-efficiency are needed. We propose a method that aims at learning a mapping from the observations into a lower-dimensional state space. This mapping is learned with unsupervised learning using loss functions shaped to incorporate prior knowledge of the environment and the task. Using the samples from the state space, the optimal policy is quickly and efficiently learned. We test the method on several mobile robot navigation tasks in a simulation environment and also on a real robot.  A video of our experiments can be found at: \url{https://youtu.be/dgWxmfSv95U}
%\hl{questions the paper addresses:}
%\begin{itemize}
%  \item how does the shaping of the loss functions used by the state-net influence the learned state representation?
%  \item how to choose the loss functions for the state-net to encode information of the task?
%    \item how to choose the state dimensionality that encodes the important info for the RL algorithm? How to choose the correct number of independent state components?
 %   \item is there any relation between loss values, PCA and negative peaks of the reward over time?
%    \item can the reward signal be used for steering the (unsupervised) learning of the state representation?
%\end{itemize}

\end{abstract}


%%%%%%%%%%%%%%%%%%%%%%%%%%%%%%%%%%%%%%%%%%%%%%%%%%%%%%%%%%%%%%%%%%%%%%%%%%%%%%%%
\section{Introduction}\label{Introduction}

Artificial intelligence (AI) is the key element to bring robots in everyday life. Robots will be asked to accomplish many different and complex tasks (e.g. navigation and exploration of unknown environments, objects manipulation and human-interaction, etc.) and theses challenges require the ability to extract meaningful information or features from the data perceived by the sensors. Because of the high task complexity, usually, multiple sensor modalities are employed. The so-called observation space, i.e. the space containing the sensory data, has a dimensionality much higher than the so-called state space, i.e. the space containing the meaningful information for solving the task.

Traditionally, this leads to complicated manual preprocessing of the data, feature engineering, and coding of the task solution. Even though very successful, feature engineering suffers from a lack of generalizability and reusability in different contexts. For each new task, it is usually necessary a new preprocessing stage and often the coding of a new solution.

Deep Reinforcement Learning (RL) \cite{Sutton1998} has been used for decision making in many different scenarios without the need for any feature engineering. RL aims at learning the mapping from the observation space to the action space directly from the data obtained through the interaction with the environment and the reward received for each action taken.  The direct end-to-end mapping from observation to action has successfully solved a huge variety of tasks \cite{mnih2015human} (e.g. videogames, robot path planning, dexterous manipulation, etc.), but it usually requires a high amount of data that are not often easy to obtain (e.g. training on real robotics hardware). Furthermore, no control over the learning of the task-relevant information is present, but the RL algorithms extract, without any supervision, the important features out of the input data.

\begin{figure*}[t!] 
	\centering
    \includegraphics[page=1,width=0.645\textwidth]{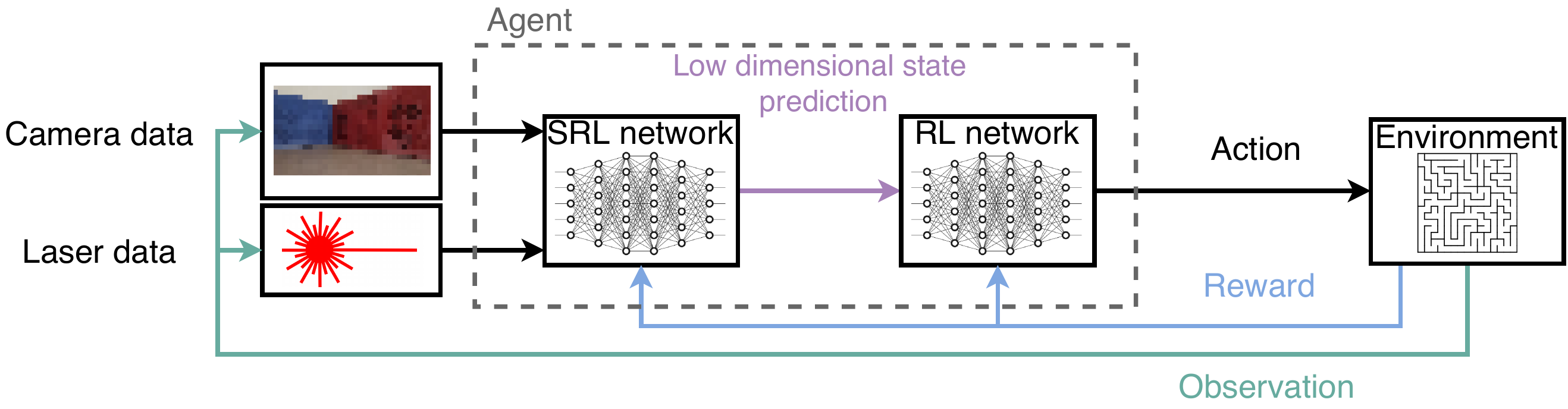}
	\captionsetup{justification=centering}
	\caption{Overall end-to-end framework combining State Representation Learning and Reinforcement Learning.}
	\label{fig:full_picture}
\end{figure*}

State Representation Learning (SRL) is the name given to the process of learning and encoding the task-meaningful information from the observation space to the so-called state space, i.e. the space containing only the task-relevant information. Usually, the state space has dimensionality much smaller than the observation space. %For RL algorithms, learning the mapping from these states to actions is much faster and sample efficient than the direct mapping from observations to actions.
The mapping from observation to states can be learned using supervised learning methods using labeled data, i.e. true value of the states. However, these are difficult and expensive to obtain.
In this work, we specifically focus on a method for tackling the state representation learning problem using unsupervised or self-supervised learning, i.e. without the use of the true value of the states. 
However, to aid the learning of a meaningful representation we use the concept of priors introduced by \cite{Bengio2013} and further developed by \cite{Jonschkowski2015}. With the priors, we model the prior knowledge about the world that can be used to inject information in the state representation learning problem. For example, it is possible to phrase these priors as loss functions for neural networks.
The authors believe that unsupervised and self-supervised methods combined with general prior knowledge of the world are the keys to achieve higher degrees of intelligence and autonomy in robotics. 

In this work, we aim at incorporating the reward function properties into the state representation learning process through the priors. We extend the concept of the priors to multiple sensor modalities (a very common scenario in robotics), to multi-targets navigation tasks and transfer learning from simulation to real robot. The general framework used is shown in Figure \ref{fig:full_picture}.

The rest of the paper is organized as follows: Section \ref{Related_work} presents the related work in the scope of this  paper,  while  Section  \ref{Theory} provides the theoretical information about RL and SRL. Then, Section \ref{Methodology} discusses the proposed methodology. Section \ref{Experiments} provides information about the experiments designed and Section \ref{Results} presents and discusses the results obtained. Section \ref{Conclusions} concludes the paper. %Section \ref{Discussion} analyses and  discusses the overall findings and Section \ref{Conclusions} concludes the paper. 

%%%%%%%%%%%%%%%%%%%%%%%%%%%%%%%%%%%%%%%%%%%%%%%%%%%%%%%%%%%%%%%%%%%%%%%%%%%%%%%%
\section{Related work}\label{Related_work}

SRL aims at learning the correct encoding of the state information out of the raw sensor observations. The quality of the state representation is crucial for decision-making, performances of RL algorithms, and their generalization capabilities. The mapping from observations to states is commonly learned with neural networks \cite{SRL_survey} using mostly auto-encoders (AEs) and variation of these (e.g. variational AEs, denoising AEs, etc.) 
Accordingly to \cite{SRL_survey}, three main methodologies can be followed to learn meaningful state representations for RL. 

The first one relies on the observation reconstruction using, for example, AEs. An AE is a neural network composed by an encoder $\phi$ that maps observations $o$ to latent state variables $s$ of lower dimensionality, i.e. $s_t=\phi(o_t)$, and a decoder $\phi^{-1}$ that reconstructs the observations from these latent variables, i.e. $\hat{o}_t=\phi^{-1}(s_t)$. Because of the imposed dimensionality reduction, the autoencoder tries to extract the relevant features from the observations in order to minimize the reconstruction error loss $L_{AE}=(o_t-\hat{o}_t)^2$. Variations of autoencoder learning are used in \cite{AE_visuomotor}, \cite{pendulum_AE}, \cite{robustness_priors} and \cite{Alvernaz_2017}.

Second, it is possible to leverage on forward models, i.e models predicting the next state $s_{t+1}$ given the current state $s_t$ and action $a_t$ and inverse models, i.e models predicting the action $a_t$ given the state $s_t$ and the next one $s_{t+1}$. Forward and inverse models are used in \cite{goroshin2015learning}, \cite{van_Hoof_2016}, \cite{jonschkowski2017pves} and \cite{agrawal2016learning}.

The third methodology, the one used in this work, uses prior knowledge about the task and the environment to shape the state space. The prior knowledge is encoded in form of loss functions used to train the neural network in charge of the observation-state mapping. To this category belongs the work proposed in  \cite{Jonschkowski2015}, \cite{robustness_priors} and \cite{morikstate}.

Independently on the chosen method, the state representation should be able to efficiently compress the observation space, with minimum information loss, to a state space with Markovian properties \cite{property_SRL}, i.e from a single state $s_t$, it is possible to choose the best action without ambiguity. The aim is to transform a Partially Observable Markov Decision Process (POMDP), in the observation space, which is difficult to solve, requires memory and high amount of sample, to a simple Markov Decision Process (MDP), in the state space, that can be efficiently solved by any RL algorithm. The state representation should be also able to generalize to unseen observations with similar features.

\section{Background}\label{Theory}

The main elements of RL \cite{Sutton1998} are the agent and the environment. The agent, by interacting with the environment, learns the mapping between state $s_t$ and action $a_t$, i.e. the policy $a_t = \pi(s_t)$, by receiving a reward $r_t$ for each action taken. The ultimate goal of the agent is to find the optimal policy, i.e the policy that maximizes the total cumulative discounted reward in Equation (\ref{total_cumulative_rew}).

\begin{equation}
R = \Sigma_{t=0}^T \gamma^{t} r_{t}
\label{total_cumulative_rew}
\end{equation}
where $\gamma$ is the discount factor and $r_t$ is the reward received by taking action $a_t$ is the state $s_t$.

Many RL algorithms estimate the state value function $V(s_t)$ or the state-action value function $Q(s_t,a_t)$ and infer the optimal policy from it. These methods are called in literature value-function-based approaches. Q-learning \cite{qlearnig} is one of them. Q-learning learns the state-action value function $Q(s_t,a_t)$, which is an estimate of how good is to choose a certain action in a given state. 

\begin{figure*}[ht] 
	\centering
    \includegraphics[page=1,width=0.55\textwidth]{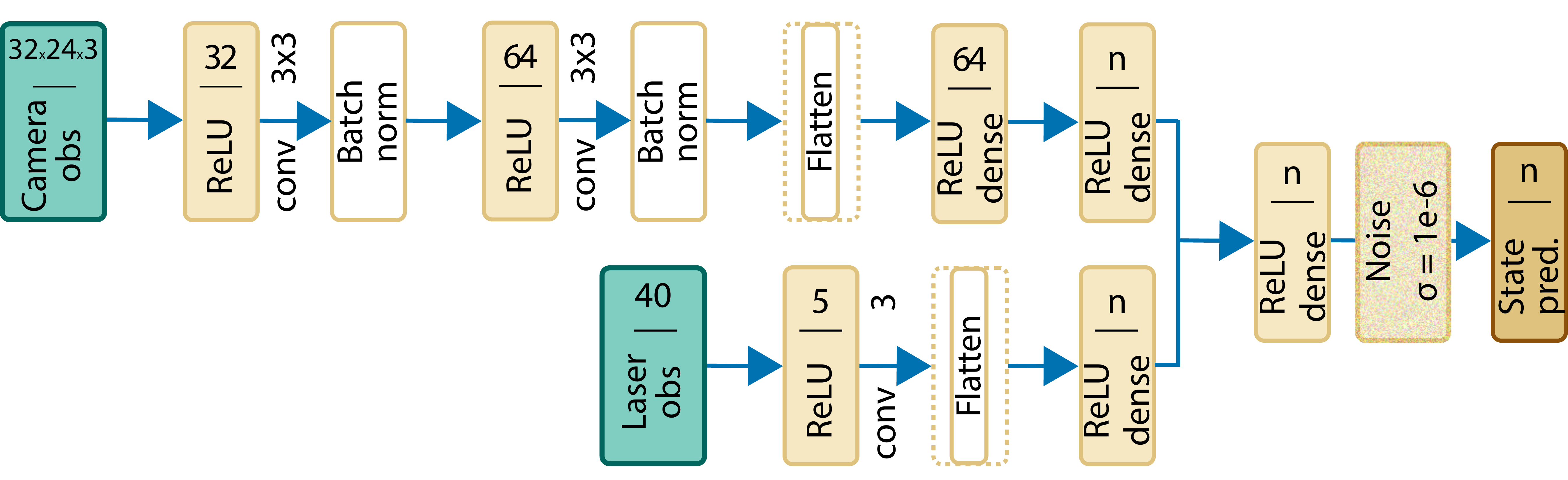}
	\captionsetup{justification=centering}
 	\caption{Architecture of the \textit{State-net}.}
	\label{fig:state-net}
\end{figure*}

Deep Q-Network (DQN) \cite{mnih2013playing}, improves the original Q-learning by approximating the state-action value function with a neural network. However, while the algorithm is now capable of handling continuous state spaces or big discrete state-action spaces (very common in many applications), the algorithm inherits the training instabilities of the neural network. When training neural networks, the first assumption is of independent and identically distributed data (i.i.d), however, in RL, the samples are collected from trajectories, thus strongly temporally correlated. This temporal correlation of the samples makes the training of the Q-network highly unstable, thus Experience Replay \cite{lin1993reinforcement} is used to break the temporal correlation between the samples as it generates training batches composed by randomly sampled data points.
The second problem is related to the loss function of the Q-network (see Equation \ref{loss_Q}). The loss requires a target $r_t+\max_{a_{t+1}}Q(s_{t+1},a_{t+1})$ to compute the temporal difference error that is then back-propagated to adjust the parameters of the network. However, this target is non-stationary and it is predicted using the same network that is updated. This generates, again, instability. To solve this issue, Double DQN (DDQN) \cite{van2016deep} uses a copy of the Q-network $Q'$ to compute this target \textit{Q}-values.

\begin{equation}
    L  =  (r_t+\gamma Q'(s_{t+1},\max_{a_{t+1}}Q(s_{t+1}, a_{t+1}))-Q(s_t,a_t))^2
    \label{loss_Q}
\end{equation}

%In many interesting application of RL in robotics, the state $s_t$ of the environment is not directly observable and has to be inferred from the high dimensional observations $o_t$ coming from the sensors. The problem of mapping, $\phi$, from observation to state, $s_t = \phi(o_t)$, is commonly addressed as SRL. 
%When learning a state representation, usually encoder networks are used, i.e. networks with output dimensionality much smaller that the input dimensionality, in order to compress the information and ultimately extract the meaningful features. 

%%%%%%%%%%%%%%%%%%%%%%%%%%%%%%%%%%%%%%%%%%%%%%%%%%%%%%%%%%%%%%%%%%%%%%%%%%%%%%%%

\section{Methodology}\label{Methodology}
%In this Section, first the proposed approach is introduced (Section \ref{overall_struct}),  followed by the analysis of the new priors (Section \ref{reward-shaped-priors}). Eventually, the neural networks architecture and the training procedure are described (Section \ref{nn_architecture}).

\subsection{Proposed approach} \label{overall_struct}

Learning useful representations of the environment is essential for autonomous robotics and decision making.  However, the mapping from the observation space, usually high-dimensional, and to the state space, usually lower-dimensional, is not straightforward. Here, we notice that with state space we intend the space of important information, necessary for learning the optimal policy for a given task using reinforcement learning. In general, the ground truth information is not always available or easy to obtain. Therefore, we aim at learning a valid state representation in an unsupervised fashion. However, we employ generic domain knowledge to shape the state representation: the robotics priors \cite{Jonschkowski2015}.
In this work, we proposed an adaptation of the original ones. The use of priors makes the learning of a state representation sample efficient and possible after a few training epochs. 

The multi-modal observations are fed to the \textit{State-net} (see Figure \ref{fig:state-net}), i.e. the network in charge of encoding the important information from the data and compressing them into a lower-dimensional state vector. The \textit{State-net} design was inspired by the architecture proposed in \cite{Babuska}.

\begin{figure}[h!] 
	\begin{center}
		\includegraphics[page=1,width=0.19\textwidth]{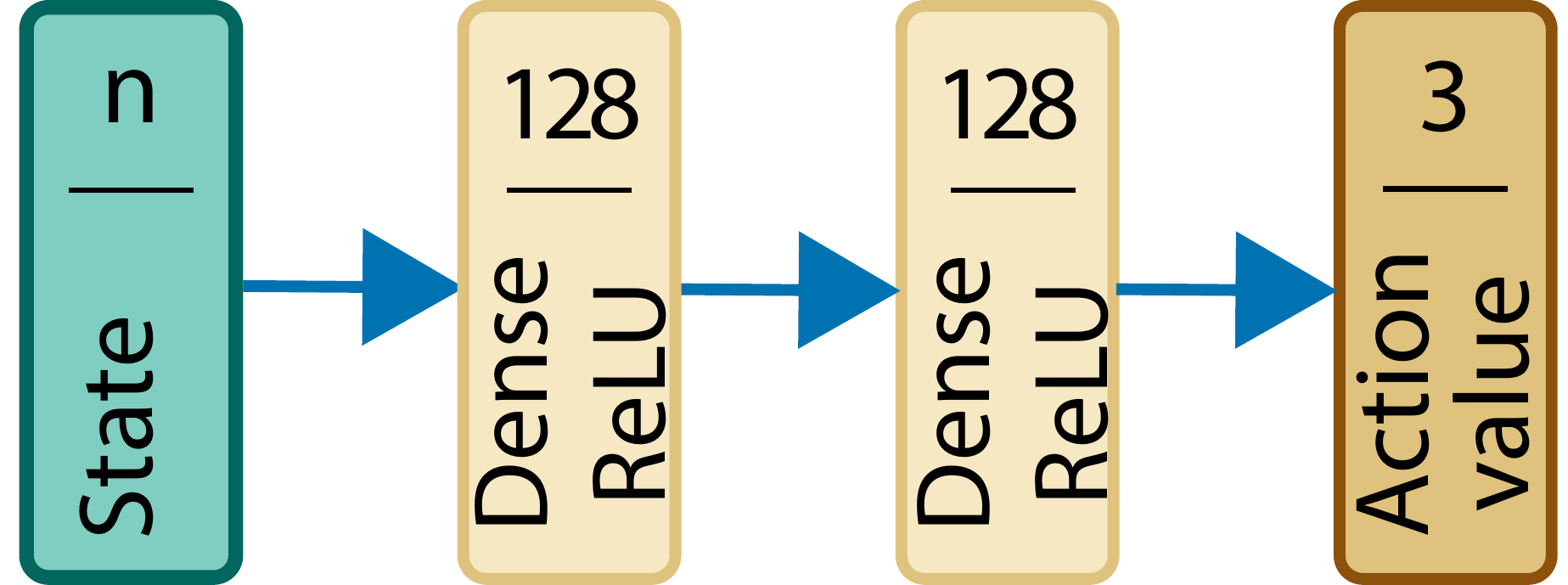}
	\end{center}
	\captionsetup{justification=centering}
	\caption{Architecture of the \textit{Q-net}.}
	\label{fig:rl-net}
\end{figure}

The state vector is then passed as input of a \textit{Q-network} (see Figure \ref{fig:rl-net}) in order to estimate the state-action value function that is then used it to choose the optimal action. DDQN was chosen for its simplicity and popularity, but the method is not dependent on this choice and any other RL algorithm can be used, both with discrete and continuous action spaces. This scheme is shown in Figure \ref{fig:full_picture}.

\begin{figure*}[!h]
\centering
\begin{subfigure}{0.18\textwidth}
  \centering
  \includegraphics[width=0.60\textwidth]{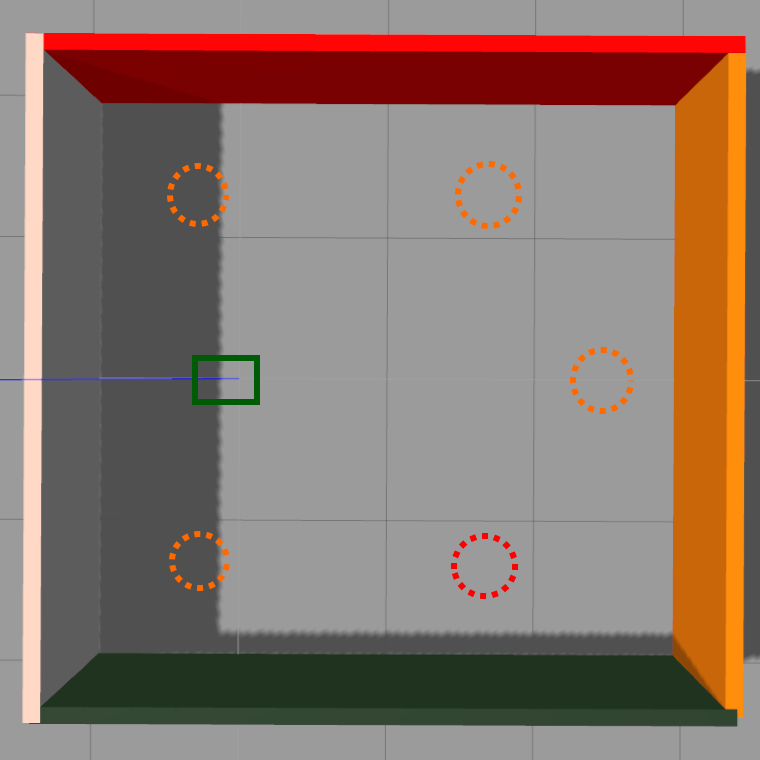}
  \captionsetup{justification=centering}
  \caption{\textit{Env-1}}
  \label{fig:env1}
\end{subfigure}
\begin{subfigure}{0.18\textwidth}
  \centering
  \includegraphics[width=0.45\textwidth]{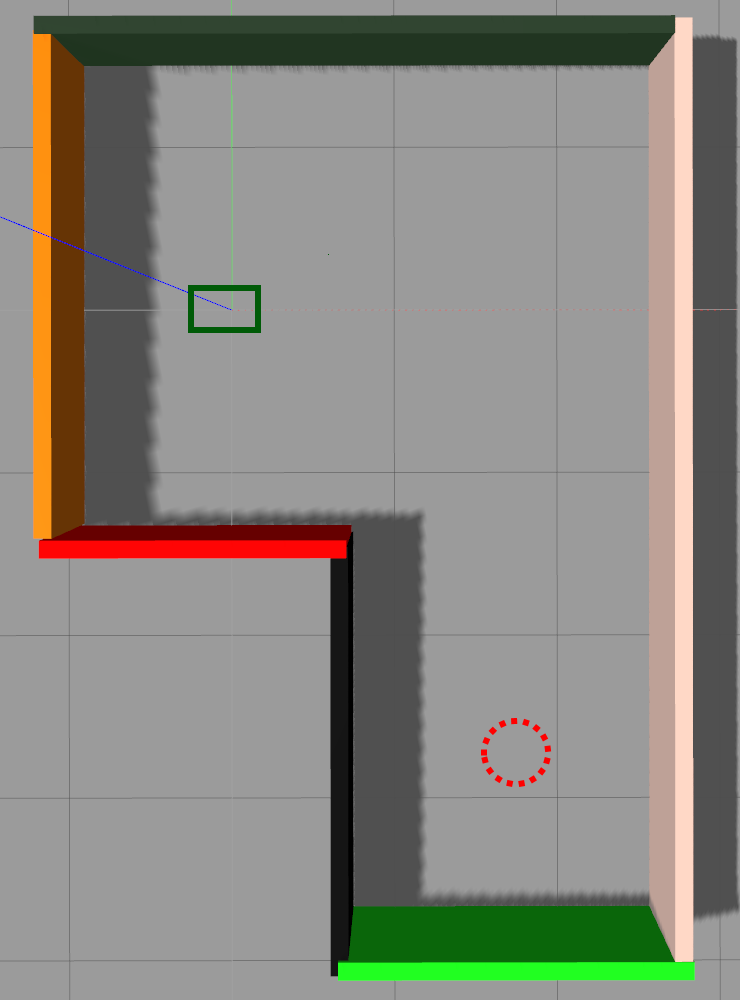}
  \captionsetup{justification=centering}
  \caption{\textit{Env-2}}
  \label{fig:env2}
  \end{subfigure}
 \begin{subfigure}{0.18\textwidth}
  \centering
  \includegraphics[width=0.60\linewidth]{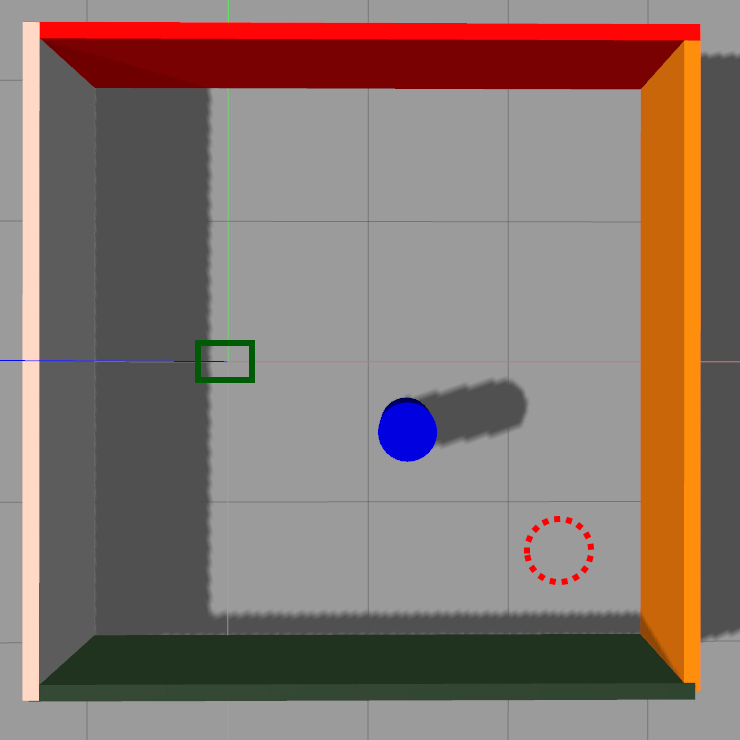}
  \captionsetup{justification=centering}
  \caption{\textit{Env-3}}
  \label{fig:env3}
  \end{subfigure}
  \begin{subfigure}{0.18\textwidth}
  \centering
  \includegraphics[width=0.60\textwidth]{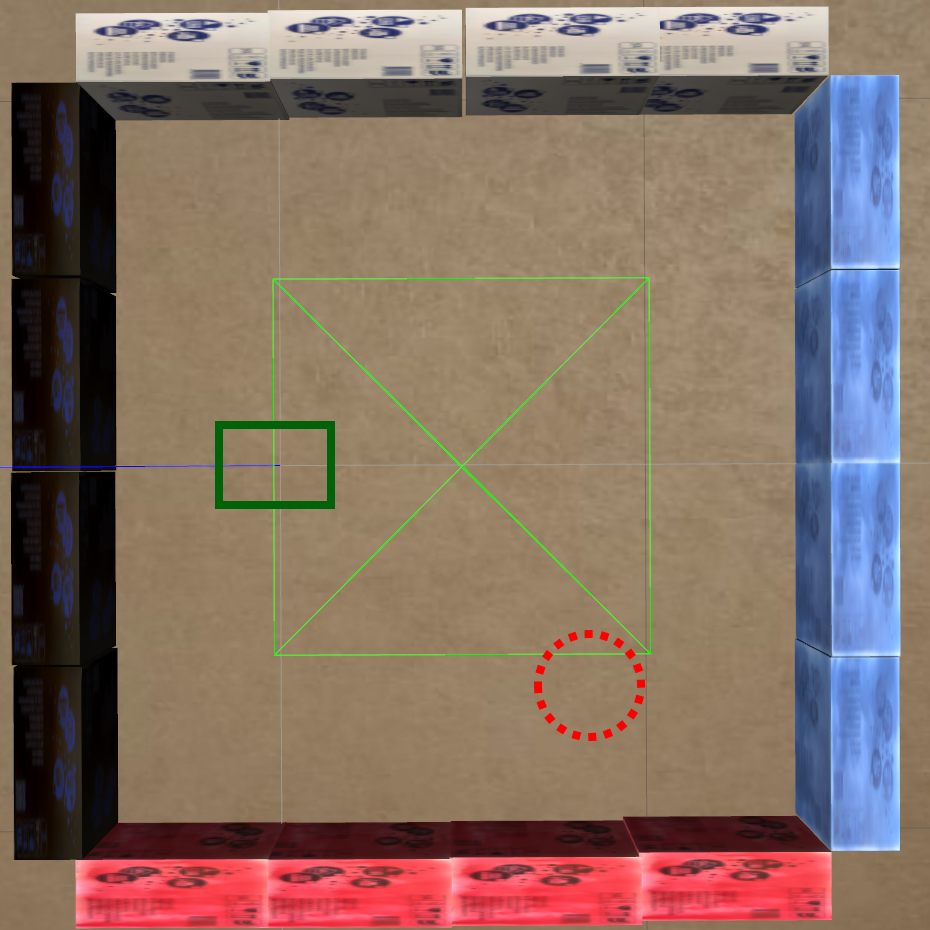}
  \captionsetup{justification=centering}
  \caption{\textit{Env-4}}
  \label{fig:env4}
  \end{subfigure}
  \begin{subfigure}{0.18\textwidth}
  \centering
  \includegraphics[width=0.60\textwidth]{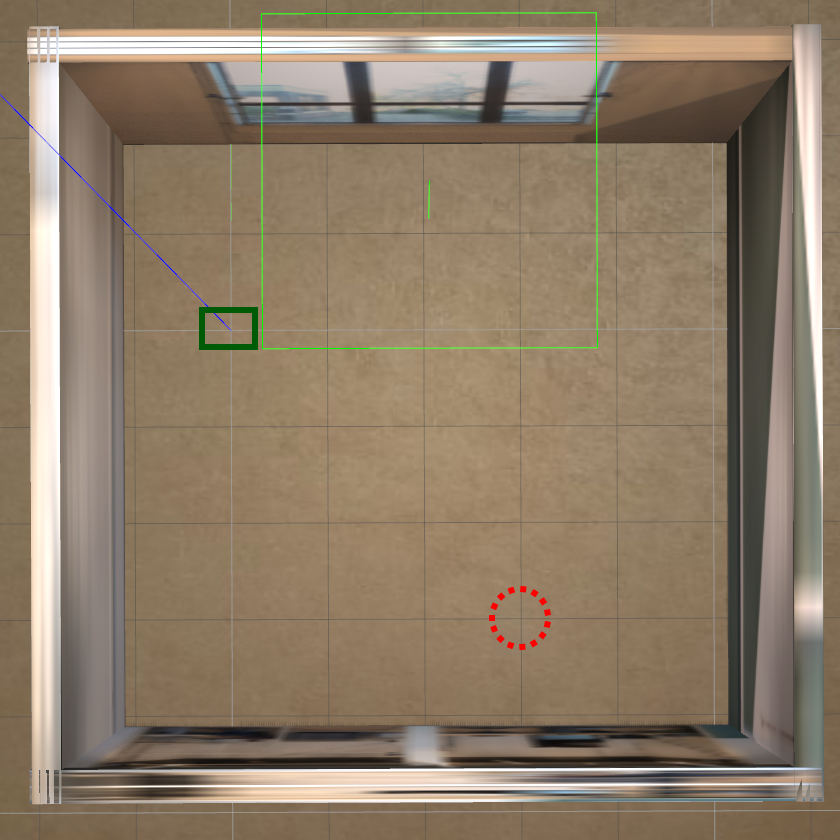}
  \captionsetup{justification=centering}
  \caption{\textit{Env-5}}
  \label{fig:env5}
  \end{subfigure}
\captionsetup{justification=centering}
\caption{Simulation environments. The robot starting position is highlighted by green rectangles, while possible targets locations by red circles.} 
\label{fig:envs}
\end{figure*}

\subsection{Reward-shaped priors}\label{reward-shaped-priors}
Our approach builds upon the priors introduced in \cite{Jonschkowski2015} and aims at addressing the following research questions:
\begin{enumerate}
    \item How can the reward function, through the priors, be used for shaping the learning of the state representation?
    \item How can the concept of priors be extended to multiple sensor modalities, different environments, and multi-targets navigation problems?
    \item To what extent, can the representation learned, using the priors, in the simulation environment be transferred effectively to the real robot without further re-training?
\end{enumerate}

The priors used in this work to train the \textit{State-net} are listed below, where $s_t$ corresponds to the state prediction given the observation $o_t$, $r_t$ to the reward achieved by taking action $a_t$ in the state $s_t$, $\Delta s_t = s_{t+1} - s_{t}$ and $\Delta r_t = r_{t+1} - r_{t}$\\
\\
\textit{Simplicity Prior}: The task-relevant information lies in a space with dimensionality much smaller than the sensory observations.
\\
\textit{Temporal Coherence}: State changes are slow and dependent only on the most recent past. This can be interpreted as an enforcement of the Markov's assumption.
\begin{equation}
    L_1 = \mathop{\mathbb{E}}\pmb{[}\mid \mid \Delta s^2_t \mid \mid\pmb{]}
\label{temp_coherence}
\end{equation}
\\
\textit{Reward Proportionality (new prior)}: Similar reward changes should induce similar state changes. These reward changes are the results of actions, but actions can be continuous or with different levels of abstraction (e.g. in the case of Hierarchical RL) and the notion of similarity is difficult to define for those cases. This new prior aims at clustering together states with similar reward variations independently on the kind of action taken.
\begin{equation}
\small{
   L_2 =\mathop{\mathbb{E}}\pmb{[}(\mid \mid \Delta s_{t_2} \mid \mid - \mid \mid  \Delta s_{t_1} \mid \mid)^2 \pmb{|} \mid \Delta {r}_{t_2} \sim \Delta {r}_{t_1}  \mid \ \pmb{]}
      }
\label{reward_proportionaliy}
\end{equation}
where $\Delta s_{t_2}$ and $\Delta s_{t_1}$ correspond to two state pairs from different time instants $t_1$ and $t_2$.
\\
\textit{Causality (new prior)}: Dissimilar rewards are a symptom of state dissimilarity. With analogous reasoning as before, the constraint to similar actions in the Causality prior of \cite{Jonschkowski2015} is removed.

\begin{equation}
\small{
L_3 = \mathop{\mathbb{E}}\pmb{[}e^{-{\mid \mid{s}_{t_2} - {s}_{t_1} \mid \mid}^2} \pmb{|} \ {r}_{t_2} \neq {r}_{t_1} \   \ \pmb{]}
}
\label{causality}
\end{equation}
\\
\textit{Reward Repeatability (new prior)}: Reinforces the similarity of states when presenting the same reward variation, not only in magnitude, but also in direction.

\begin{equation}
\small{
L_4 = \mathop{\mathbb{E}}\pmb{[}e^{-{\mid \mid{s}_{t_2} - {s}_{t_1} \mid \mid}^2}(\mid \mid \Delta s_{t_2} -   \Delta s_{t_1} \mid \mid)^2\pmb{|} \mid \Delta {r}_{t_2} \sim \Delta {r}_{t_1} \mid \  \pmb{]}
}
\label{repreatability}
\end{equation}

The overall loss function, Equation (\ref{total_loss}), used for training the \textit{State-Net} is equal to the weighted sum of the different priors with the addition of L2-regularization term.

\begin{equation}
{
\mathcal{L} = \omega_1L_1 + \omega_2L_2 + \omega_3L_3 + \omega_4L_4 + \omega_5L_{reg}
}
\label{total_loss}
\end{equation}

The weights of the single loss functions (in Equation (\ref{total_loss})) were chosen equal to $\omega_1=3$, $\omega_2=15$, $\omega_3=15$, $\omega_4=15$ and $\omega_5=3$ to balance the contribution of the single loss functions. This combination gave good empirical results, but no optimization procedure was run to find the best set of weights and only grid-search was employed.

In RL, the reward function is defined and shaped based on task-specific knowledge to allow the agent to learn optimal behaviors. However, in the context of SRL, a task cannot be efficiently learned if an informative representation hasn't been learned yet. We believe that the best representation is the one that incorporates meaningful information for solving the task, therefore it shouldn't be learned independently from the chosen reward function. The new priors (\ref{reward_proportionaliy}), (\ref{causality}), and (\ref{repreatability}) were developed to achieve this goal: shaping the state representation using not only the environment observations, but also the rewards. In particular, the reward variation from two states is used to further impose the Markov's assumption, during the observation compression step. Ideally, we would like to obtain a regularized state space that is Markovian, i.e. a standard RL algorithm by looking at a single state prediction can choose the optimal action without the need for any memory structure.

\subsection{Neural networks architecture and training regime} \label{nn_architecture}

The \textit{State-Net}, Figure \ref{fig:state-net}, is an encoder network, i.e. neural network with output dimensionality much smaller than the input dimensionality. The samples from the two sensor modalities are passed through two separate network branches and they are both used to make to independent state predictions of dimension \textit{n}. The two predictions are concatenated and then fed through a final fully connected (dense) layer that combines them to produce the final state prediction, again of dimension \textit{n}. The considerations on choice of the state dimensionality are shown in Section \ref{Results}.
The state predictions and the actions are then used to estimate the \hbox{\textit{Q}-values} using the neural network shown in Figure \ref{fig:rl-net}

As shown in \cite{Babuska}, the state representation network shouldn't be updated with the same frequency of the reinforcement learning network due to the generation of high learning instability. Therefore, we normally train the \textit{Q-Net} (Figure \ref{fig:rl-net}) at each training step, while we update only after a fixed number of training episodes the \textit{State-Net} (\ref{fig:state-net}). The frequency of the update of the \textit{State-Net} is chosen to be a  trade-off between training too often and generate instability and training too rarely and slow down the learning of the optimal policy. The optimal policy cannot be learned without an informative state representation.
In the episodes right after the updates of the \textit{State-Net}, the rewards achieved by the RL-agent may drop due to the sudden changes of the state representation. To avoid learning local optimal policies, we hold constant the value of the $\epsilon$ of the $\epsilon$-greedy exploration policy of DDQN.

\begin{figure*}[ht!]
\centering
\begin{subfigure}{0.30\textwidth}
  \includegraphics[width=\textwidth]{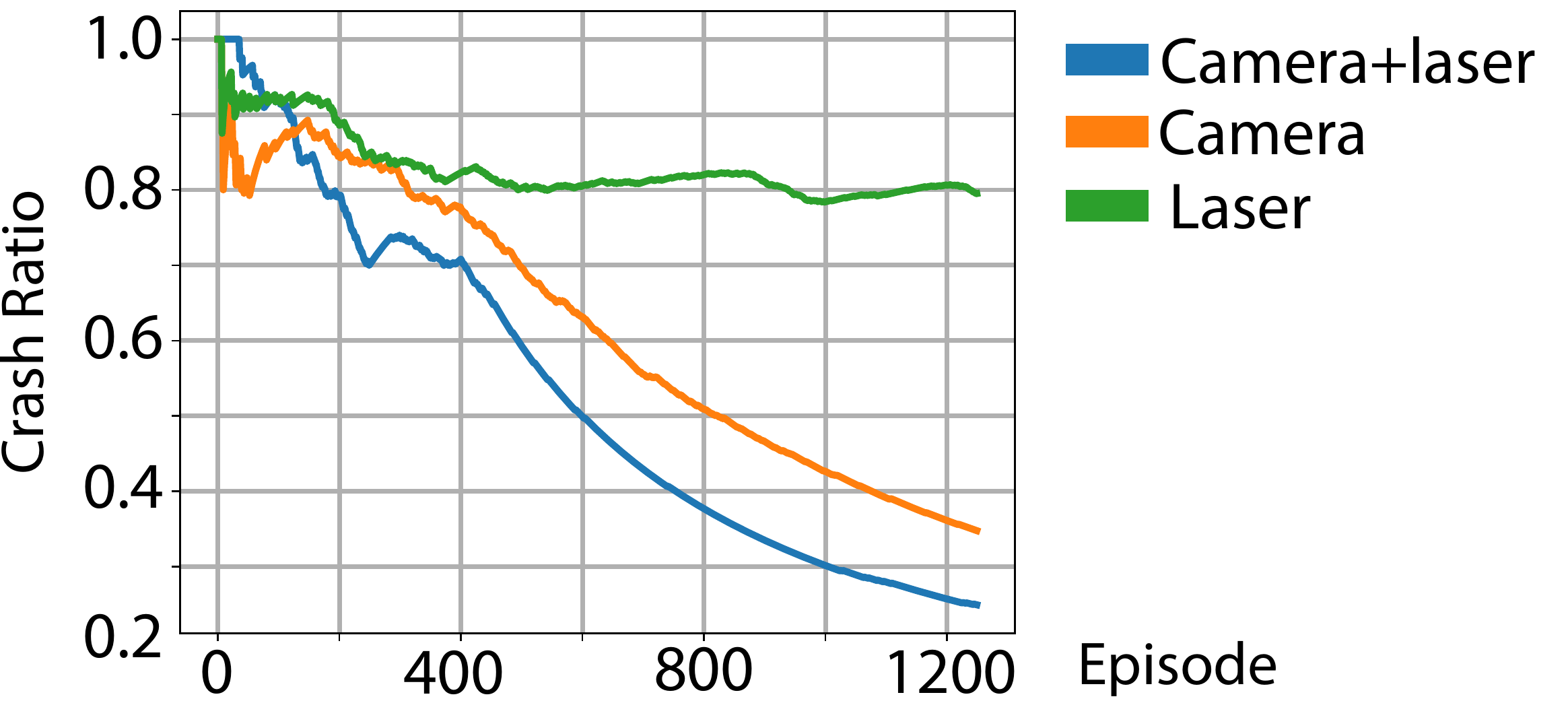}
  \captionsetup{justification=centering}
  \caption{}
  \label{fig:crash_diff_sensors}
\end{subfigure}
\begin{subfigure}{0.31\textwidth}
    \includegraphics[width=\textwidth]{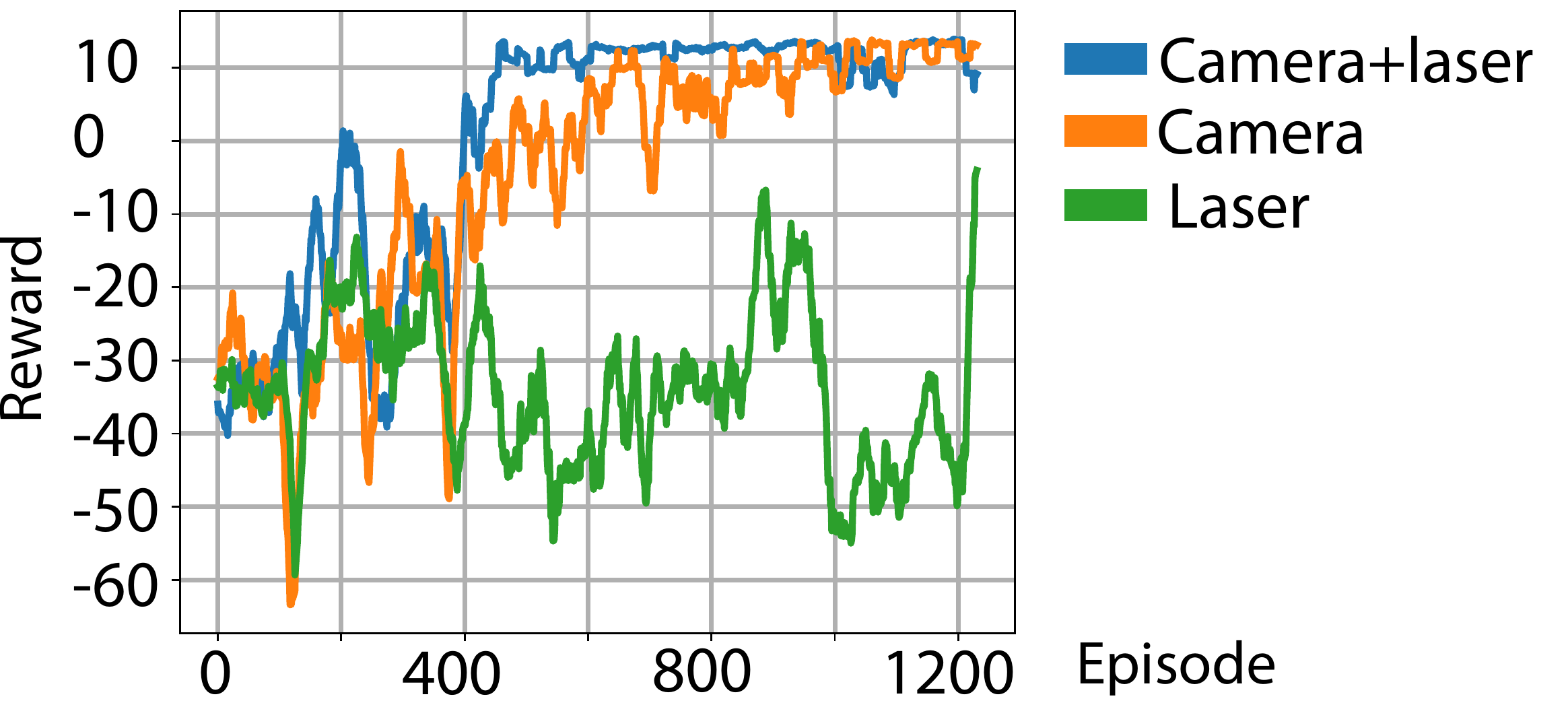}
    \captionsetup{justification=centering}
    \caption{}
    \label{fig:rew_diff_sensors}
\end{subfigure}
\captionsetup{justification=centering}
\caption{Crash ratio (\ref{fig:crash_diff_sensors}) and cumulative reward (\ref{fig:rew_diff_sensors}) obtained during training using different sensor modalities.} 
\label{fig:crash_rew_diff_sensors}
\end{figure*}

%To stabilize the training, we propose the use of a network with structure equal to the \textit{State-Net}, similar to the siamese architecture proposed in \cite{robustness_priors}. However, in this case, the second network parameters are copied from the \textit{State-Net}, but at the previous update epoch. The \textit{Memory-Net} (Figure \ref{fig:memory-net}) outputs a state of the same dimension of the \textit{State-Net} and these two state predictions are concatenated and sent as input of the \textit{Q-Net}. This allows to the \textit{Q-Net} to learn the changes in the state predictions before the update of the \textit{State-Net} to the new state predictions after the update.

%\begin{figure}[h!] 
%	\begin{center}
%		\includegraphics[width=0.50\textwidth]{memory_net.png}
%	\end{center}
%	\captionsetup{justification=centering}
%	\caption{\textit{Memory-net} architecture}
%	\label{fig:memory-net}
%\end{figure}

%%%%%%%%%%%%%%%%%%%%%%%%%%%%%%%%%%%%%%%%%%%%%%%%%%%%%%%%%%%%%%%%%%%%%%%%%%%%%%%%

\section{Experiments design}\label{Experiments}
%In this Section the experiments are presented. The setup is presented in Section \ref{setup}, then the RL parameters choices are presented in  Section \ref{rl_settings}. The simulation environments are described in Section \ref{sim_experiments}, followed by the multi-targets experiments in Section \ref{multi-targets}. Eventually, the transfer learning experiments from simulation to real robot are discussed in Section \ref{transfer_learning}.

\subsection{Mobile robot navigation with camera and LIDAR} \label{setup}
When autonomously navigating, mobile robots are usually equipped with multiple sensors (sensor modalities) in order to be able to gather the highest amount of information from the environment. Commonly used sensors for perceiving the world are cameras and laser range scanners (LiDARs). Therefore, we equipped our robot (Turtlebot 3 waffle) with a camera (FOV 60 degrees) and a 2D LiDAR (FOV 360 degrees). The approach is first tested in the ROS-Gazebo 3D simulation environment and later evaluated on the real robot (again Turtlebot 3 waffle).

\subsection{Reinforcement learning algorithm settings}\label{rl_settings}
The algorithm chosen is DDQN with inputs the state predictions from the \textit{State-Net} and with output the \textit{Q}-values, one estimate per action. The agent can choose among 3 discrete actions: respectively, go forward, turn right, and turn left. To study the effect of different reward functions on the state representation learned with the new set of priors (Equation (\ref{temp_coherence})-(\ref{repreatability})), two different reward functions are tested:
\begin{itemize}
    \item Reward function based on the distance between the robot and the target (Equation (\ref{distance_rew}))
    \item Reward function based on the orientation of the robot with respect to the target (Equation (\ref{orientation_rew}))
\end{itemize}

\begin{equation}
    \small
    r(s_t)=\begin{cases}
    r_{\text{reached}}, & d \leq d_{min},\\
    r_{\text{crashed}}, & s_{ts},  \\
    1-e^{\eta_1 d}, & \text{otherwise}.
    \end{cases}
    \label{distance_rew}
\end{equation}

\begin{equation}
    \small
    r(s_t)=\begin{cases}
    r_{\text{reached}}, & d \leq d_{min},\\
    r_{\text{crashed}}, & s_{ts},  \\
    1-e^{\eta_2 \theta}, & \text{otherwise}.
    \end{cases}
    \label{orientation_rew}
\end{equation}

where $d$ is the distance of the robot to the target, estimated using the odometry information, $\theta$ is the robot orientation with respect to the target, $d_{min}$ is the minimum distance threshold below which the navigation target is considered reached and $\eta_1$ and $\eta_2$ are scaling factors for the exponential functions. $r_{reached}$ and $r_{crashed}$ are respectively a bonus for reaching the target and a penalty for hitting an obstacle, i.e. a terminal state $s_{ts}$. These two reward functions are a common choice for solving navigation tasks.

\subsection{Navigation tasks in different environments}\label{sim_experiments}We first compare the new priors with the ones from \cite{Jonschkowski2015} in order to highlight similarities and differences in the environment in Figure \ref{fig:env1}. We then analyze the choice of the state dimensionality as being a crucial aspect of the RL performances. Furthermore, we study, through t-SNE \cite{tsne}, PCA \cite{PCA} and correlation analysis, if the \textit{State-Net} trained with the priors (Equation (\ref{temp_coherence})-(\ref{repreatability}) succeeds in encoding the meaningful information for solving the navigation task. In the case of the mobile robot navigation proposed, this information corresponds to the physical properties of the world as, for example, the pose of the robot ($x$-position, $y$-position, and $\theta$ orientation) and its distance to the target. 
Eventually, we test our approach in environments with different topologies and features (e.g. different colors of the wall, textures, etc.), shown in Figure \ref{fig:env2}-\ref{fig:env5}, to validate the method. We also again study the information encoded by that \textit{State-Net} and to what extent these are dependent on the environment shape.

\subsection{Multi-targets state representation} \label{multi-targets}
We perform experiments to assess the priors in case of a more complicated task: learning a state representation for multiple navigation targets. During training at every episode, a target is sampled from a uniform distribution. We slightly adapt the observation vector to include the location, ($x$,$y$) coordinates, of the target. This information is directly passed to the last dense layer of the \textit{State-Net}. 

\subsection{Transfer learning experiments} \label{transfer_learning}
Transfer learning is an important element for deploying RL algorithms on real robots, but it is usually limited by the simulation-reality gap, i.e. the difference that always exists between simulation and the real world. However, if informative high-level features are extracted from the observations, the RL policies, trained on these features, gain robustness and can be transferred from simulation to real without any undesirable training on the real robots.

%%%%%%%%%%%%%%%%%%%%%%%%%%%%%%%%%%%%%%%%%%%%%%%%%%%%%%%%%%%%%%%%%%%%%%%%%%%%%%%%

\section{Results and Discussion}\label{Results}

%In this Section the results are presented. First, the effect of the two sensor modalities is studied in Section \ref{results_diff_sens_mod} and simulation results in the different environments are shown in Section \ref{results_4walls}. Then, the results of the multi-targets navigation task are discussed in Section \ref{results_multitarget}, followed by the transfer learning experiments on the real robot in Section \ref{results_real_robot}. 

The state predictions are analyzed using Principal Components Analysis (PCA) \cite{PCA} and t-Distributed Stochastic Neighbor Embedding (t-SNE) \cite{tsne}. These two techniques for dimensionality reduction allow us to visualize high dimensional datasets, understand and explain the learned state representation.

\subsection{Mobile robot navigation with camera and LiDAR} \label{results_diff_sens_mod}
 Here, we analyze the influence of the different sensor modalities on the learned representation. In particular, we compare the quality of learned representation through the crash ratio and the total cumulative reward when camera and LiDAR are used (Figure \ref{fig:crash_diff_sensors} and \ref{fig:rew_diff_sensors}), only the camera is used (Figure \ref{fig:crash_diff_sensors} and \ref{fig:rew_diff_sensors}) and only the LiDAR is used (Figure \ref{fig:crash_diff_sensors} and \ref{fig:rew_diff_sensors}).
%\begin{itemize}
%    \item camera and LiDAR are used (Figure \ref{fig:crash_diff_sensors} and %\ref{fig:rew_diff_sensors})
%    \item only the camera is used (Figure \ref{fig:crash_diff_sensors} and \ref{fig:rew_diff_sensors})
%    \item only the LiDAR is used (Figure \ref{fig:crash_diff_sensors} and \ref{fig:rew_diff_sensors})
%\end{itemize}
%When both sensor modalities are used, Figure \ref{fig:camera_lidar}, the clustering, assessed by looking at the smoothness of the color gradient, of the state prediction is improved with respect to camera and laser only. Furthermore,
When both sensors are employed, the crash ratio is reduced (see Figure \ref{fig:crash_diff_sensors}) and the convergence speed is improved (see Figure \ref{fig:rew_diff_sensors}). This shows how the representation learned with the priors is capable of fusing the different sensor modalities to obtain the best set of features.

\subsection{Navigation tasks in different environments}\label{results_4walls}

%The results related to the comparison with the original priors introduced in \cite{Jonschkowski2015} are presented (Section \textit{Comparison with original priors}). Then, 
%the analysis of the state dimensionality (Section \textit{Analysis of the state dimensionality}) and of what it is really encoded in the state using the new priors (Section \textit{Reward-shaped state representation}) is shown.

\subsubsection{Comparison with original priors} \label{comparison_old_priors}
\begin{figure}[t]
    \centering
    \includegraphics[page=1,width=0.3\textwidth]{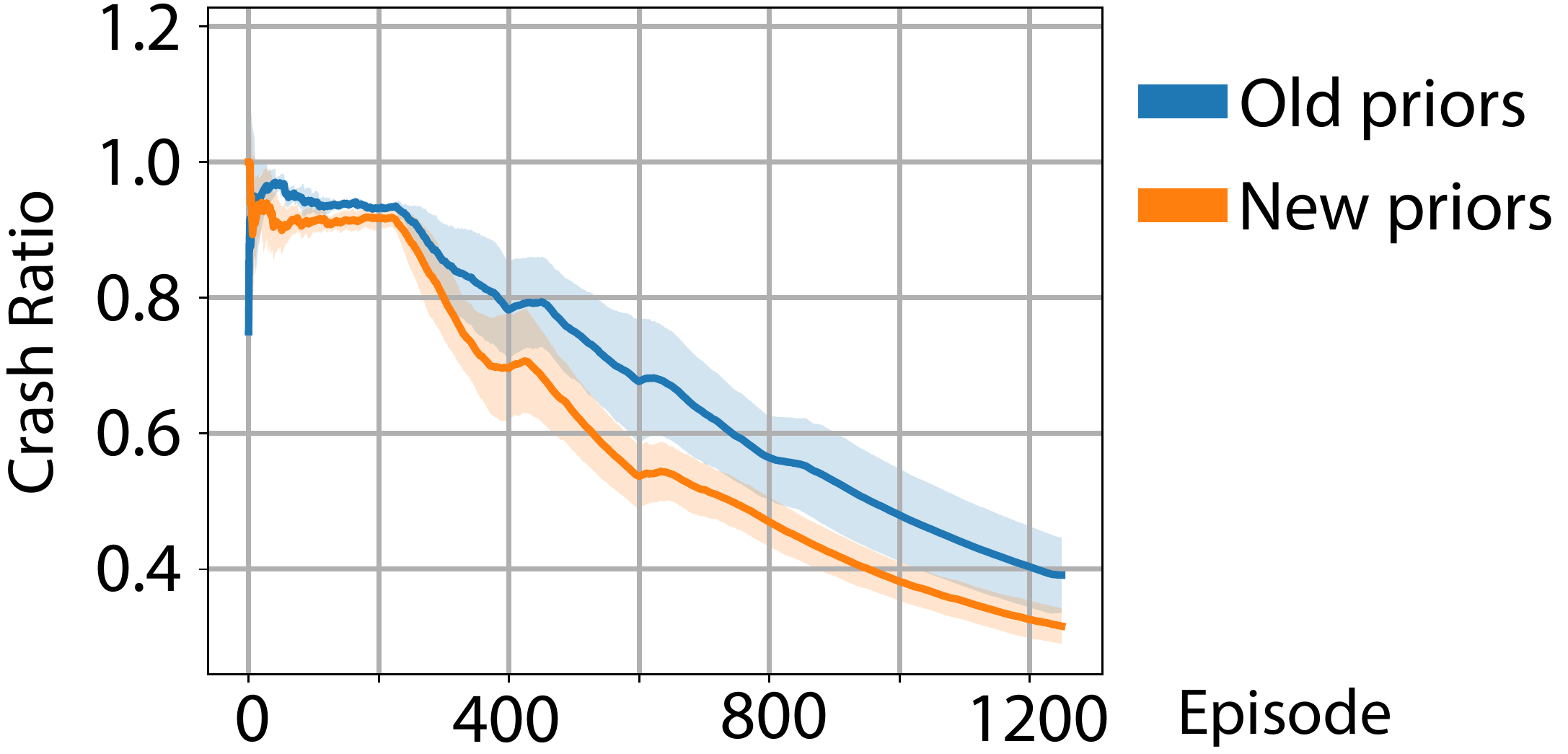}
    \caption{Crash ratio when the new priors and the original priors are used on the same navigation task.}
    \label{fig:comp_priors}
\end{figure}

The comparison with the priors, introduced in \cite{Jonschkowski2015}, is done by comparing the effect of the different learned state representation on the performance of the RL-agent in the environment depicted in Figure \ref{fig:env1}. For the sake of a fair comparison, the training and testing environment is similar to the one used in \cite{Jonschkowski2015}. Furthermore, the same neural networks and hyperparameters are used. In Figure \ref{fig:comp_priors}, the crash ratio during training is shown when the proposed priors and the original priors are used.

As shown in Figure \ref{fig:comp_priors}, the new set of priors is able to improve the RL performances by reducing the average crashing ratio and its variance during training. This is because the new priors enforce a state representation that incorporates the properties of the reward function that has to be maximized by the RL algorithm. This is further elaborated on in Section \ref{results_rew_shaped_rep}.

\subsubsection{Analysis of the state dimensionality} \label{results_state_dim_analysis}
\begin{figure}[h!]
    \centering
	\includegraphics[page=1,width=0.3\textwidth]{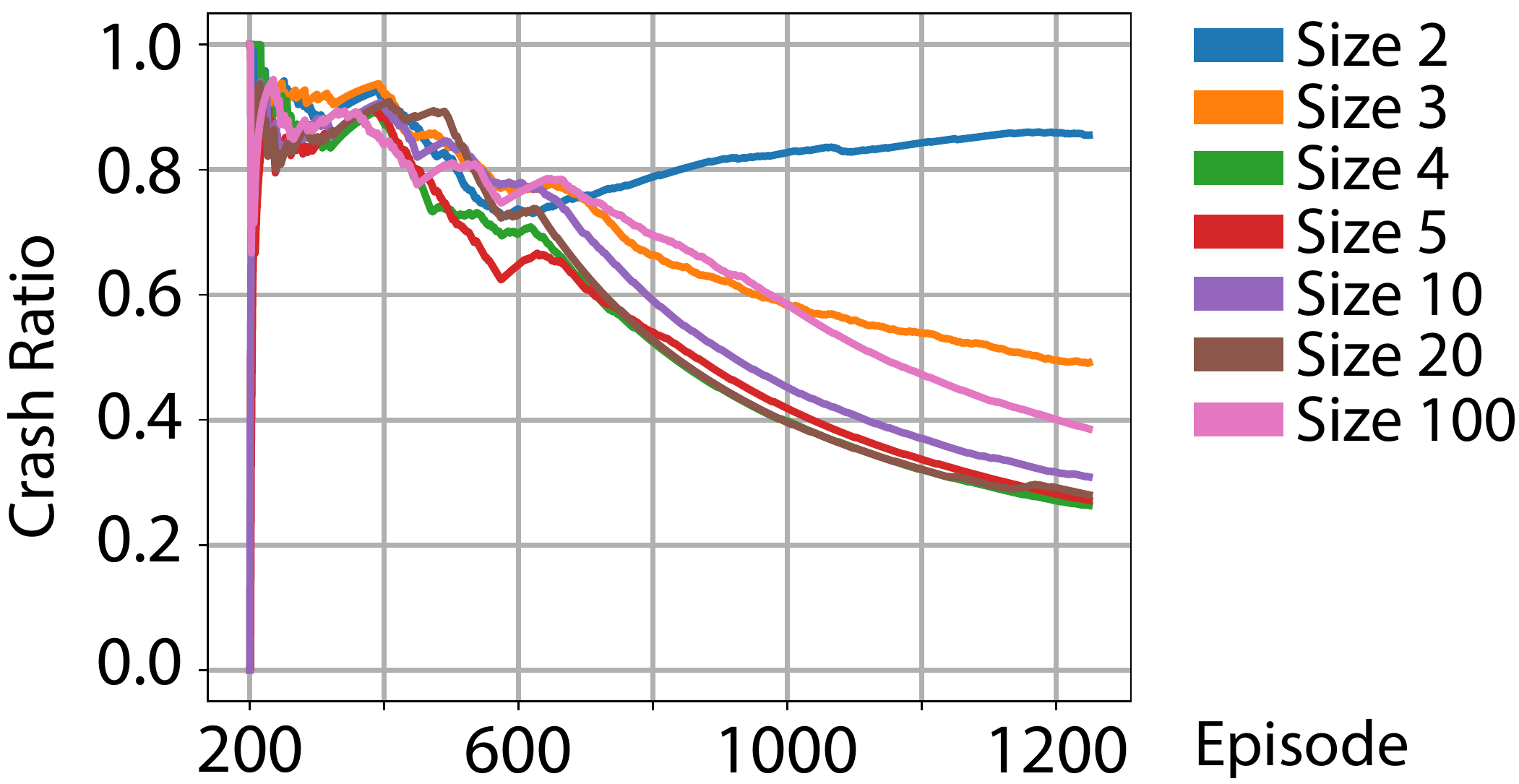}
	\captionsetup{justification=centering}
	\caption{Crash ratio of the agent with respect to the state dimension.}
	\label{fig:crash_state_dim}
\end{figure}

The choice of the state dimensionality is crucial for RL performances. To test it, we analyze the crash ratio, i.e. the number of times an episode ends due to a collision with an obstacle over the total number of episodes, in relation to the choice of the state dimension. This choice corresponds to the choice of the output dimension of the \textit{State-Net}. The results are shown in Figure \ref{fig:crash_state_dim}. It is possible to notice that if the state dimension is chosen too small with respect to the optimal one, the encoding step loses much important information due to the exaggerated compression. This is the case for the state dimension equal to 2 and 3. In those cases, the RL-agent struggles to reduce the collisions and improve the policy. On the other hand, if the dimension is chosen too big, for example, equal to 100, the performances of the RL-agent are slowed down due to the lack of compression and the curse of dimensionality. The RL-agent has to learn which information has to be ignored.

\begin{figure}[t]
    \centering
	\includegraphics[page=1,width=0.33\textwidth]{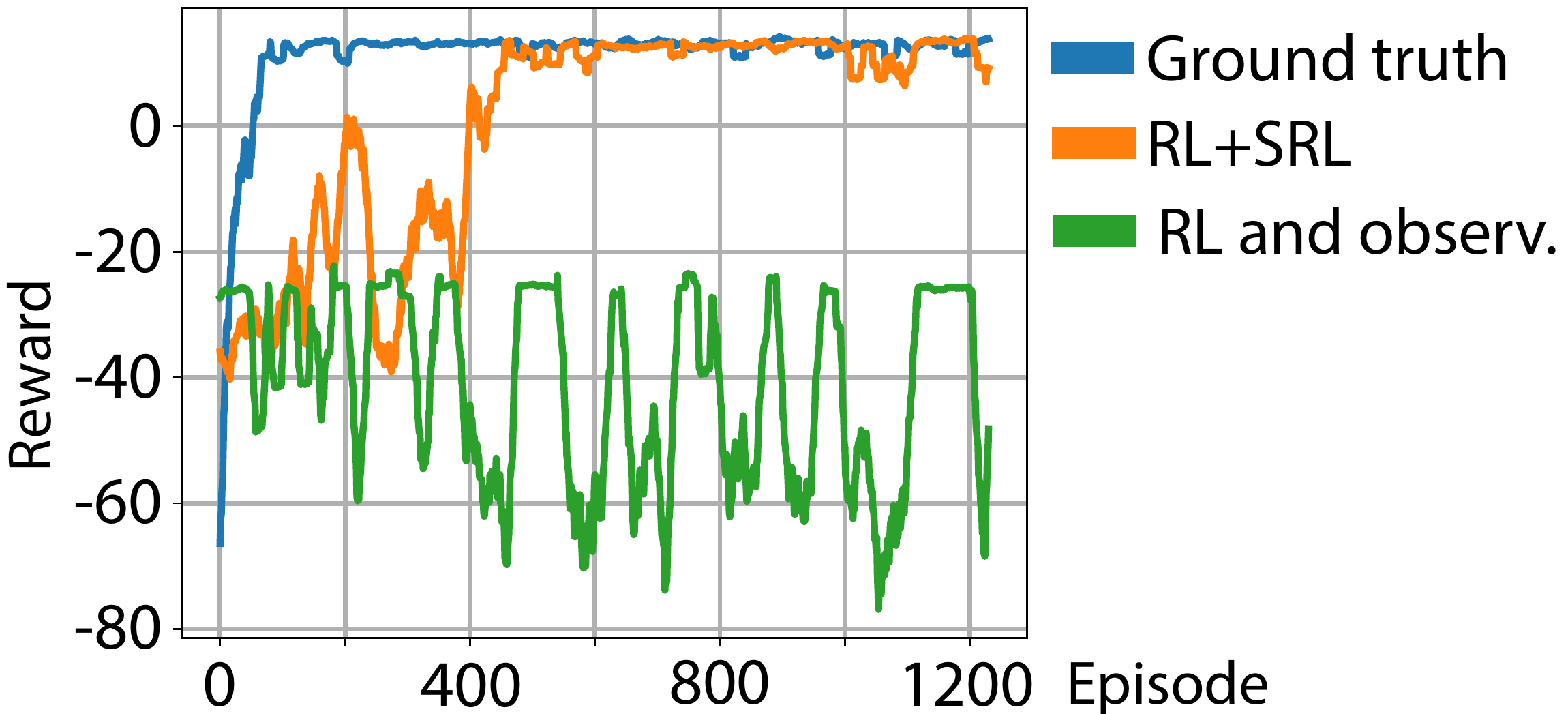}
	\caption{Comparison of RL based on true pose, SRL combined with RL (ours) and RL with input the raw observations.}
	\label{fig:comparison_RLGT-SRLRL-obsRL}
\end{figure}

We compare our approach with RL using the true pose of the robot ($x$-position, $y$-position and $\theta$ orientation) and end-to-end RL based on observations (see Figure \ref{fig:comparison_RLGT-SRLRL-obsRL}) in the environment in Figure \ref{fig:env1}. As expected, RL based on the ground truth quickly converges to the optimal solution (blue line in Figure \ref{fig:comparison_RLGT-SRLRL-obsRL}), however the knowledge of the ground truth is a limiting factor in many real-world scenarios. When the state representation is combined with RL  (orange line in Figure \ref{fig:comparison_RLGT-SRLRL-obsRL}) after few updates of the \textit{State-Net} (occurring at episode 200 and 400 respectively), the policy converges to the optimal solution with slope very similar to the policy using the ground truth. The policy directly based on observation (green line in Figure \ref{fig:comparison_RLGT-SRLRL-obsRL}) cannot converge in the time window of 1200 episodes. This result proves the effectiveness of the state representation learning using the priors.

% \begin{figure*}[h!]
% \centering
% \begin{subfigure}{0.23\textwidth}
%   \centering
%   \includegraphics[width=1.0\linewidth]{pic/distance_rew_distance_clustering.png}
%   \captionsetup{justification=centering}
%   \caption{}
%   \label{fig:dist_clustering}
% \end{subfigure}
% \begin{subfigure}{0.23\textwidth}
%   \centering
%   \includegraphics[width=1.0\linewidth]{pic/distance_rew_orientation_clustering.png}
%   \captionsetup{justification=centering}
%   \caption{}
%   \label{fig:orient_clustering}
%   \end{subfigure}
%   \begin{subfigure}{0.23\textwidth}
%   \centering
%   \includegraphics[width=1.0\linewidth]{pic/orientation_rew_distance_clustering.png}
%   \captionsetup{justification=centering}
%   \caption{}
%   \label{fig:dist_clustering_o}
% \end{subfigure}
% \begin{subfigure}{0.23\textwidth}
%   \centering
%   \includegraphics[width=1.0\linewidth]{pic/orientation_rew_orientation_clustering.png}
%   \captionsetup{justification=centering}
%   \caption{}
%   \label{fig:orient_clustering_o}
%   \end{subfigure}
% \captionsetup{justification=centering}
% \caption{Distance-based reward (\ref{fig:dist_clustering} and \ref{fig:orient_clustering}) and orientation-based reward (\ref{fig:dist_clustering_o} and \ref{fig:orient_clustering_o}). Distance vs orientation clustering (t-SNE visualization)} 
% \label{fig:dist_rew}
% \end{figure*}

Through PCA, we study the actual dimensionality of the encoded state space by counting the number of uncorrelated components. The method is tested for all the environments in Figure \ref{fig:envs}. The results obtained are shown in Table \ref{env_comparison}. 

The state representation learned with the priors is not dependent on the topology of the environment (e.g. its shape) or the choice of the features (e.g. wall's colors or textures) as the number of uncorrelated component is consistently 4 in \textit{Env-1}, \textit{Env-2}, \textit{Env-4} and \textit{Env-5} (see Table \ref{env_comparison}). This proves that the state representation learning method proposed generalizes well in different environments. Interestingly, in \textit{Env-3}, when an obstacle is present on the optimal trajectory towards the target, the state representation can encode that information. This is reflected in the number of uncorrelated components, as a fifth one emerges. This again proves that the state representation learned with the new priors can encode the task-relevant information.

\begin{table} [h!]
\centering
\caption{Different environment results.}
\scalebox{0.75}{
\begin{tabular}{ |c|c|c| } 
\hline
 Environment & \textit{State-Net} output dim & Nr. uncorr. components \\
\hline
%\multirow{3}{4em}{$\alpha$} & cell2 \\ 
\textit{Env-1} & 10 & 4 \\ 
\textit{Env-2} & 10 & 4 \\ 
\textit{Env-3} & 10 & 5 \\
\textit{Env-4} & 10 & 4 \\
\textit{Env-5} & 10 & 4 \\
\hline
\end{tabular}}

\label{env_comparison}
\end{table}

In order to understand what kind of information the \textit{State-Net} encodes in the state space, we compare samples from the different principal components with the physical important properties required in any navigation task: pose of the robot ($x$, $y$, $\theta$) and distance to the target. The results of the correlation analysis, for the environment in Figure \ref{fig:env1}, are shown in Table \ref{tab_1}. A correlation exists between the real physical properties of the world and the encoded properties by the \textit{State-Net}. It is worth to mention that we are not enforcing any explicit disentanglement and uncorrelation of the state properties, as we are still in an unsupervised learning framework.
%Figure \ref{fig:PCA1_dist} and \ref{fig:PCA3_orie} show the strong correlation between the real physical properties of the world and the encoded properties by the \textit{State-Net}. It is worth to mention that we are not enforcing any explicit disentanglement and uncorrelation of the state properties, as we are still in an unsupervised learning framework.

\begin{table} [h!]
\centering
\caption{Correlation analysis of the principal components and the physical properties.}
\scalebox{0.75}{
\begin{tabular}{ |c|c|c|c|c| } 
\hline
 & x-position & y-position & orientation & distance to target \\
\hline
%\multirow{3}{4em}{$\alpha$} & cell2 \\ 
Principal component 1 & 0.86 & 0.24 & 0.18 & -0.14  \\ 
Principal component 2 & -0.28 & 0.68 & 0.7 & 0.8  \\ 
Principal component 3 & -0.32 & -0.17 & -0.37 & 0.22  \\
Principal component 4 & -0.1 & 0.19 & -0.09 & 0.13  \\
\hline
\end{tabular}}

\label{tab_1}
\end{table}

%\begin{figure*}
%\centering
%\begin{subfigure}{0.2\textwidth}
 % \centering
%  \includegraphics[width=1\linewidth]{PCA1_dist.png}
%  \captionsetup{justification=centering}
%  \caption{}
%  \label{fig:pca1}
%\end{subfigure}
%\begin{subfigure}{0.2\textwidth}
%%  \centering
%  \includegraphics[width=1\linewidth]{PCA3_orient.png}
%  \captionsetup{justification=centering}
%  \caption{}
%  \label{fig:pca3}
%  \end{subfigure}
%\captionsetup{justification=centering}
%\caption{Comparison of samples from different principal components and %true position and orientation samples} 
%\label{fig:pca_analysis}
%\end{figure*}

%\begin{figure}
%    \centering
%    \includegraphics[width=0.6\linewidth]{pic/PCA1_dist.png}{}
%    \caption{Comparison of samples from different principal components and true %position and orientation samples}
%    \label{fig:PCA1_dist}
%\end{figure}

%\begin{figure}
%    \centering
%includegraphics[width=0.6\linewidth]{pic/PCA3_orient.png}{}
%    \caption{Comparison of samples from different principal components and true %position and orientation samples}
%    \label{fig:PCA3_orie}
%\end{figure}

%\hl{change table with on with target that requires more turning}

\subsubsection{Reward-shaped state representation}\label{results_rew_shaped_rep}
\begin{figure}[!ht]
    \centering
    \includegraphics[page=1,width=0.85\linewidth]{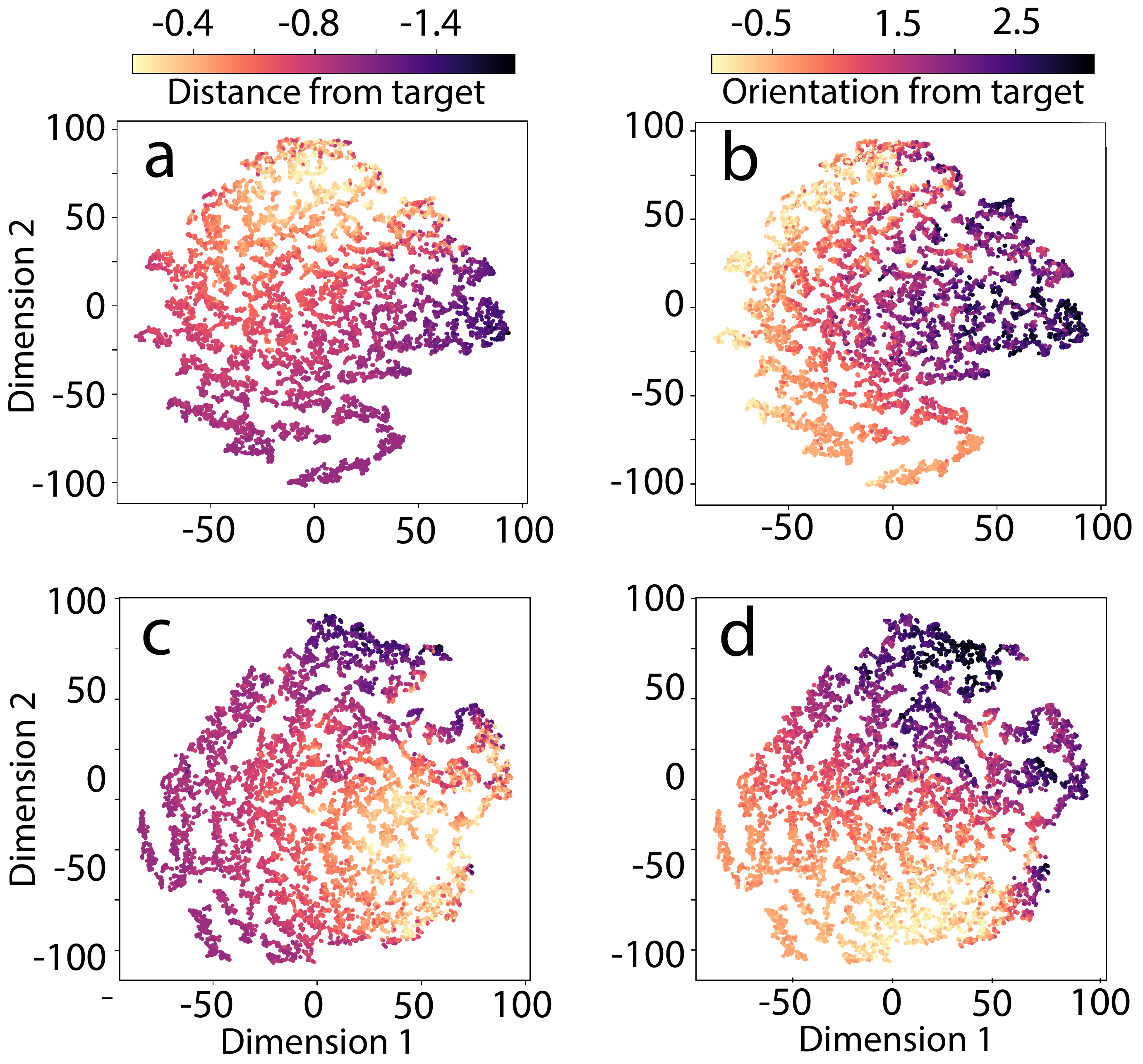}
\captionsetup{justification=centering}
\caption{Distance-based reward (a and b) and orientation-based reward (c and d). Distance vs orientation clustering (t-SNE visualization).} 
\label{fig:clustering_rew}
\end{figure}

To test if the reward signal, combined with the new priors, can be used to effectively shape the state representation by encoding from the sensors information task-specific knowledge, we analyzed, using t-SNE, the state representations obtained when the different reward functions, in Equation (\ref{distance_rew}) and (\ref{orientation_rew}), are employed. In particular, we analyze if the clustering of the state predictions is related to the chosen reward function. In Figure \ref{fig:clustering_rew}, the clustering of the state predictions, when the reward function in Equation (\ref{distance_rew}) is used, with respect to the true distance from the target (see Figure \ref{fig:clustering_rew}a) and orientation from the target (see Figure \ref{fig:clustering_rew}b) is shown. It is possible to notice that when the reward function in Equation (\ref{distance_rew}) is used, the state representation is able to encode and cluster close together the predictions that have similar rewards, i.e similar distance from the target. When the same predictions are overlapped with the true orientation (see Figure \ref{fig:clustering_rew}b), the clustering is less effective and the prediction samples with similar orientation are spread over larger areas. This is expected since the state representation is not trained to cluster the predictions with respect to the orientation.
Analogously, the clustering of the state predictions when the reward function in Equation (\ref{orientation_rew}) is used, with respect to the true distance from the target (see Figure \ref{fig:clustering_rew}c) and orientation (see Figure \ref{fig:clustering_rew}d) is shown. When the reward function in Equation (\ref{orientation_rew}) is used, the predictions are correctly clustered with respect to the true orientation (see Figure \ref{fig:clustering_rew}d), but also with respect to the true distance (Figure \ref{fig:clustering_rew}c). This is due to the fact that the orientation with respect to the target is computed using the distance from the target along the x and y-axis, thus it is not completely independent on the distance.
These results prove that the state representation encodes task-relevant knowledge through the reward information.

\subsection{Multi-targets navigation} \label{results_multitarget}
\begin{figure*}[ht!]
\centering
\begin{subfigure}{0.33\textwidth}   
  \centering
  \includegraphics[page=1,width=0.72\linewidth]{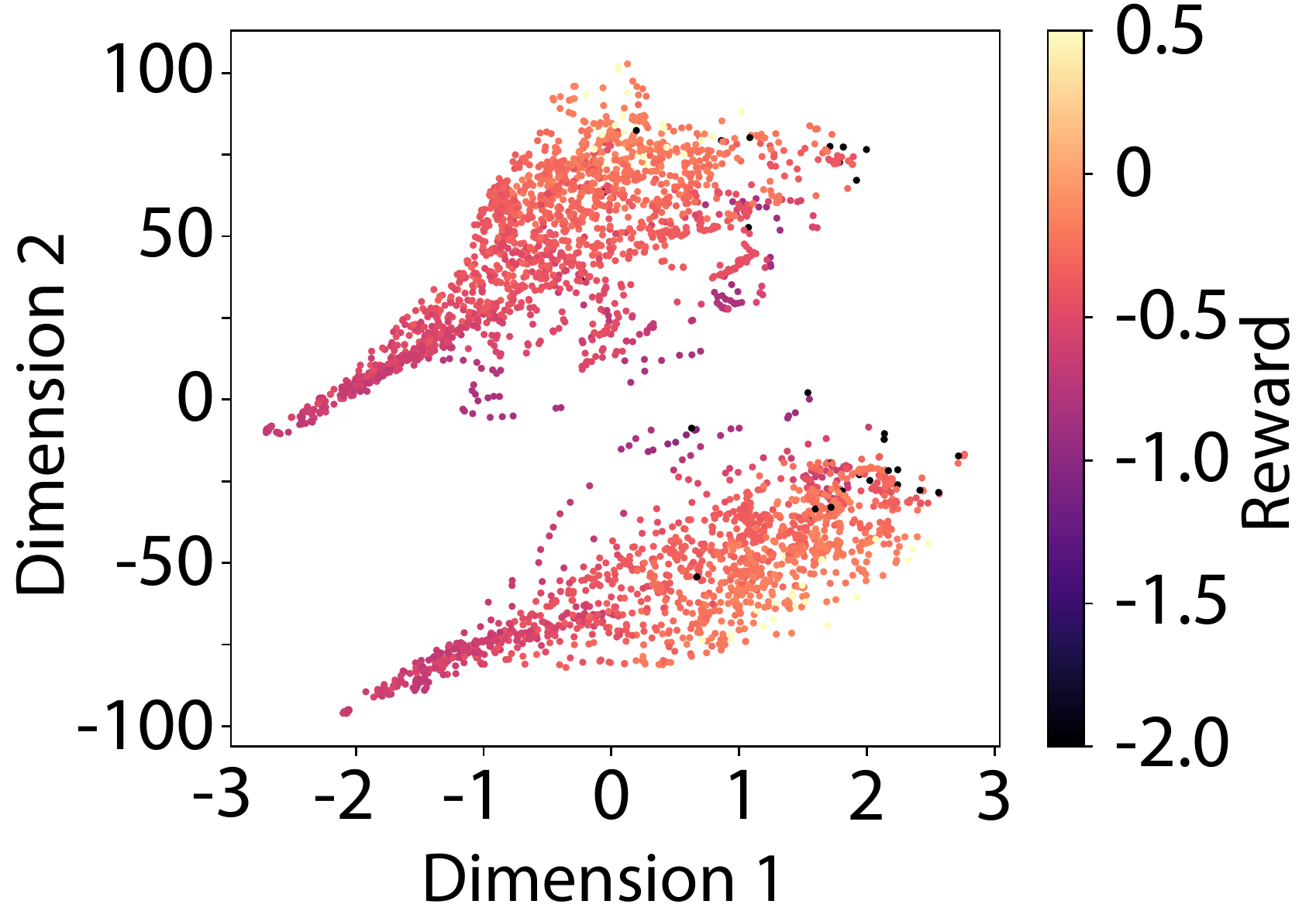}
  \captionsetup{justification=centering}
  \caption{}
  \label{fig:PCA_2targets}
\end{subfigure}
\begin{subfigure}{0.33\textwidth}
  \centering
  \includegraphics[page=1,width=0.72\linewidth]{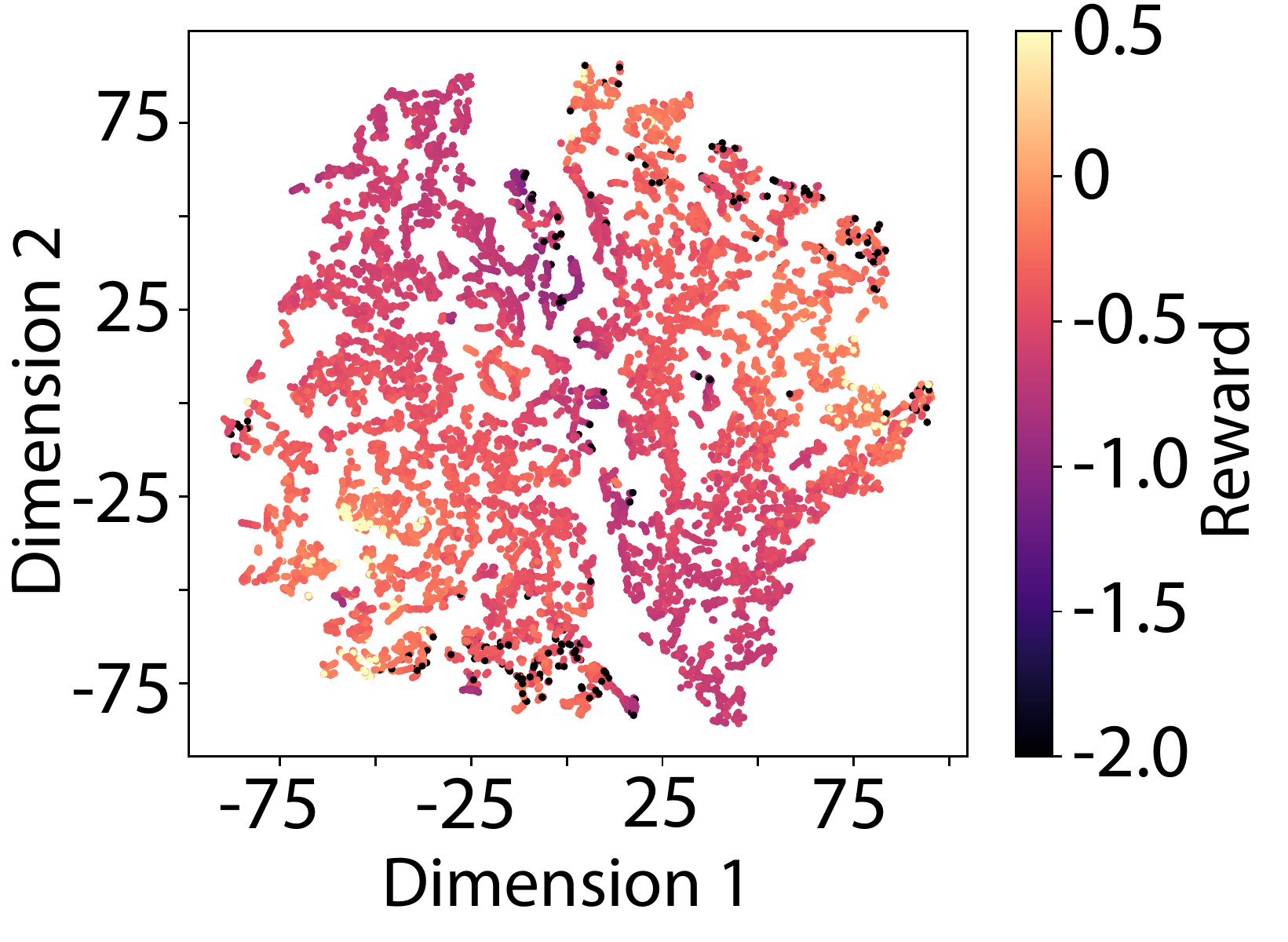}
  \captionsetup{justification=centering}
  \caption{}
  \label{fig:TSNE_2targets}
  \end{subfigure}
\captionsetup{justification=centering}
\caption{State representation learned for two different targets analysed with PCA (\ref{fig:PCA_2targets}) and t-SNE (Figure \ref{fig:TSNE_2targets}).} 
\label{fig:srl_diff_targets}
\end{figure*}

In this Section, we present the results related to multi-target navigation. In particular, we analyze if the priors are suitable for learning a state representation that is capable of differentiating between multiple navigation targets (two in this case). The results are presented in Figure \ref{fig:srl_diff_targets}, where the state predictions are analysed using PCA (Figure \ref{fig:PCA_2targets}) and t-SNE (Figure \ref{fig:TSNE_2targets}) . The state representation learned can effectively incorporate the information of the different targets and it can cluster not only in terms of the reward in Equation (\ref{distance_rew}) (this can be noticed by looking at the smoothness of the color gradient in the Figures), but also with respect to the two targets (a clear division of the state samples).

\subsection{Experiments in realistic simulation environment and on real robots}\label{results_real_robot}
\begin{figure}[!h]
    \centering
    \includegraphics[page=1,width=0.83\linewidth]{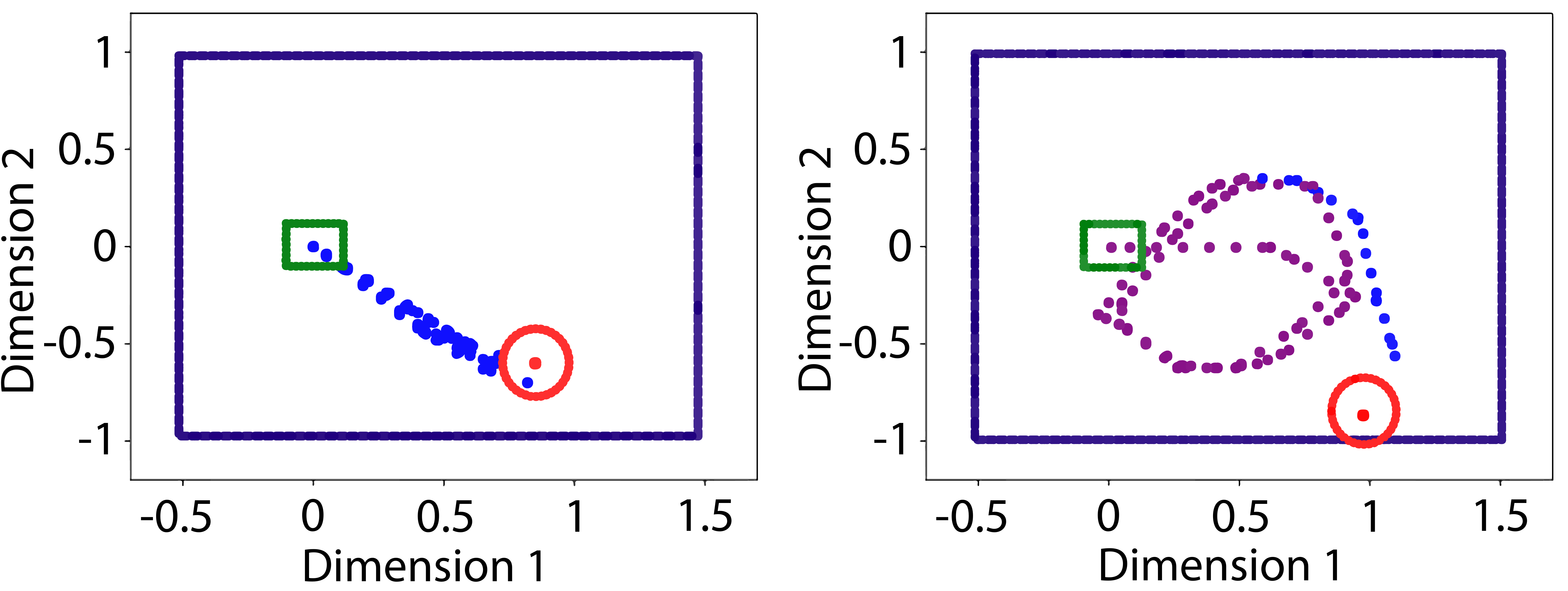}
\captionsetup{justification=centering}
\caption{Trajectories of the real robot (x-y plane) when the light is on (left). Trajectories of the real robot when the light is off (right) indicated with purple dots and when the light is turned on again. The green rectangle corresponds to the starting location of the robot, the red circle corresponds to the target location. }
    \label{fig:real_robot_trajectories}
\end{figure}

In this Section, the transfer learning experiments are presented. In particular, we show, for a single navigation target, the trajectories followed by the real robot after transferring the state representation and the policy learned in the simulation environment \ref{fig:env4}. The trajectories followed on 10 different experiments, are shown in Figure \ref{fig:real_robot_trajectories} (left).

To assess the robustness of the state representation and the policy learned in simulation to variations in the sensor reading, during the experiments we switched off the lights of the room and after few seconds we switched then back on. In Figure \ref{fig:real_robot_trajectories}, the trajectories obtained are shown.
%\begin{figure}[h!] 
%	\begin{center}
%		\includegraphics[width=0.25\textwidth]{pic/re%al_robot_traj_light_off.png}
%	\end{center}
%	\captionsetup{justification=centering}
%	\caption{Real robot trajectories when the light %is off (purple dots) and when the light is turned on %again (blue dots)}
%	\label{fig:real_robot_lights_off}
%\end{figure}
When the lights are off, the agent receives images from the camera which are very different from the one it has been trained on, thus it cannot immediately find the key features to reach the target. However, the agent doesn't perform random actions that would bring the robot to crash against an obstacle (purple dots in Figure \ref{fig:real_robot_trajectories} (right)). The agent starts a searching behavior as it rotates around in search of the correct features. Once the light is turned on again (blue dots in Figure \ref{fig:real_robot_trajectories} (right)), the agent quickly recognize the features and drives safely to the target. This can be interpreted as proof that the policy has learned robust obstacle avoidance and navigation skills.  

By extracting the meaningful features from the sensor data, not only the RL-agent learns the policy faster, but we can mitigate the simulation to reality gap and we can directly transfer the knowledge learned in the simulation environment to the real robot without any further training on the real hardware.

%%%%%%%%%%%%%%%%%%%%%%%%%%%%%%%%%%%%%%%%%%%%%%%%%%%%%%%%%%%%%%%%%%%%%%%%%%%%%%%%
%\section{Discussion}\label{Discussion}

%\hl{To be continued...}

%%%%%%%%%%%%%%%%%%%%%%%%%%%%%%%%%%%%%%%%%%%%%%%%%%%%%%%%%%%%%%%%%%%%%%%%%%%%%%%%
%%%%%%%%%%%%%%%%%%%%%%%%%%%%%%%%%%%%%%%%%%%%%%%%%%%%%%%%%%%%%%%%%%%%%%%%%%%%%%%%

\section{Conclusions}\label{Conclusions}

This paper proposes a new approach for the unsupervised learning of state representations for reinforcement learning. The state representation is learned using a new set of auxiliary loss functions, i.e. the priors. These priors are shaped using the reward function as means to incorporate the task-relevant knowledge in the state representation.
%In reinforcement learning, the reward function is defined and shaped based on task-specific knowledge to allow the agent to learn an optimal behaviour. However, a task cannot be efficiently learned if an informative state representation hasn't been learned yet. The best representation is the one that incorporates the meaningful information for solving the task, therefore it shouldn't be learned independently from the chosen reward function. 
From the tests on the different environments, the state representation, built using the reward-shaped priors, can encode the important physical properties for solving different navigation tasks. Furthermore, the state representation learned is not dependent on the topology of the environment or the textures in it. The number of uncorrelated components in the state is consistently 4. However, when an extra constraint is added in the environment (e.g. obstacles, see Figure \ref{fig:env3}), the state representation grows an extra uncorrelated component to encode information about the obstacle. The same happens in the case of multi-targets navigation tasks.
Furthermore, the priors allow the fusion of different sensor modalities (camera and LiDAR in this case).
Eventually, the state representation and policy learned in the simulation environment are successfully transferred to the real robot without further retraining.

In the future studies, we will conduct further experiments on comparing the proposed learning method with other state-of-the-art learning algorithms and especially the with auto-encoder-based approaches.
%As future work, we will compare the proposed approach with auto-encoder-based methods for state representation learning. Moreover, to further investigate the concept of reward-shaped state representation, we will study the effect of more complicated reward functions on the learned representations.

%%%%%%%%%%%%%%%%%%%%%%%%%%%%%%%%%%%%%%%%%%%%%%%%%%%%%%%%%%%%%%%%%%%%%%%%%%%%%%%%

%%%%%%%%%%%%%%%%%%%%%%%%%%%%%%%%%%%%%%%%%%%%%%%%%%%%%%%%%%%%%%%%%%%%%%%%%%%%%%%%

\section*{Acknowledgment}\label{Acknoledgment}

The authors thank Johan Engelen for the great support in the initial phase of this work, Han Wopereis for the precious help with ROS and Gazebo simulations and the reviewers for the comments that greatly improved the manuscript.

%%%%%%%%%%%%%%%%%%%%%%%%%%%%%%%%%%%%%%%%%%%%%%%%%%%%%%%%%%%%%%%%%%%%%%%%%%%%%%%%

%\bibliographystyle{unsrt} 
%\bibliography{SRL_NB_2019}   % name your BibTeX data base

%\end{document}

\bibliographystyle{IEEEtran}
\bibliography{IEEEexample}

%%%%%%%%%%%%%%%%%%%%%%%%%%%%%%%%%%%%%%%%%%%%%%%%%%%%%%%%%%%%%%%%%%%%%%%%%%
%\section*{Appendix}\label{Appendix}

%\begin{figure*}
%\centering
%\begin{subfigure}{0.25\textwidth}
%  \centering
%  \includegraphics[width=1.0\linewidth]{pic/camera+LiDAR.png}
%  \captionsetup{justification=centering}
%  \caption{}
%  \label{fig:camera_lidar}
%\end{subfigure}
%\begin{subfigure}{0.25\textwidth}
%  \centering
%  \includegraphics[width=1.0\linewidth]{pic/camera_only.png}
%  \captionsetup{justification=centering}
%  \caption{}
%  \label{fig:camera_only}
%  \end{subfigure}
 % \begin{subfigure}{0.25\textwidth}
%  \centering
%  \includegraphics[width=1.0\linewidth]{pic/LiDAR_only.png}
%  \captionsetup{justification=centering}
%  \caption{}
%  \label{fig:lidar_only}
%  \end{subfigure}
%\captionsetup{justification=centering}
%\caption{State representation with different sensor modalities (t-SNE %visualization). Camera and LiDAR (\ref{fig:camera_lidar}), camera only %(\ref{fig:camera_only}) and LiDAR only (\ref{fig:lidar_only}). } 
%\label{fig:sensor_modalities}
%\end{figure*}

\end{document}